\documentclass[9pt]{article}

\usepackage{amsfonts}

\usepackage{amsmath}

\usepackage{bm}

\usepackage{caption}

\usepackage{float}

\usepackage{graphicx}

\usepackage{soul}

\usepackage{spconf}

\usepackage[x11names]{xcolor}

\usepackage{tikz}

\newcommand{\plus}{\texttt{+}}
\newcommand{\minus}{\texttt{-}}

\newcommand{\JSD}{\text{D}_\text{JS}}
\newcommand{\KLD}{\text{D}_\text{KL}}

\DeclareRobustCommand{\cvdots}{
	\vcenter{%
		\baselineskip=4pt 
		\hbox{.}\hbox{.}\hbox{.}
}}

\usetikzlibrary{arrows.meta,
			    calc,
				decorations,
				decorations.pathmorphing,
				decorations.pathreplacing,
				positioning,
			    shapes.geometric}

\def \enTitle{Selective Sampling and Mixture Models in Generative Adversarial Networks}

\def \paperKeywords
{
	Typological analysis,
	clustering,
	selective sampling,
	multi-generator adversarial networks
}

\def \paperAbstract{
	In this paper, we propose a multi-generator extension to the adversarial training framework, in which the objective of each generator is to represent a unique component of a target mixture distribution. In the training phase, the generators cooperate to represent, as a mixture, the target distribution while maintaining distinct manifolds. As opposed to traditional generative models, inference from a particular generator after training resembles selective sampling from a unique component in the target distribution. We demonstrate the feasibility of the proposed architecture both analytically and with basic Multi-Layer Perceptron (MLP) models trained on the MNIST dataset.
}

\title{\enTitle}
\name{Karim Said Barsim,
	Lirong Yang, and
	Bin Yang \thanks{}}
\address{University of Stuttgart \\
	Institute of Signal Processing and System Theory \\
	Pfaffenwaldring 47, 70459 Stuttgart, Germany}

\begin{document}

\maketitle

\begin{abstract}\paperAbstract\end{abstract}


\section{Introduction}
\label{sec:introduction}


Over the last few years, a promising generative modeling architecture, referred to as \underline{G}enerative \underline{A}dversarial \underline{N}etwork (GAN) \cite{Goodfellow_2014_GAN}, has attracted a wide interest due to their impressive success in various application domains, including text-to-image generation \cite{Reed_2016_Text2Image}, compressive sensing \cite{Quan_2017_RefineGAN_CS_MRI}, image-to-image translation \cite{Isola_2016_Pix2Pix}, multi- and super-resolution image synthesis \cite{Denton_2015_LaplacianPyramids}, video generation \cite{Vondrick_2016_GeneratingVideos}, human driver modeling \cite{Kuefler_2017_GAIL_DriverGAN} in addition to a variety of  unsupervised and semi-supervised learning tasks \cite{Shrivastava_2016_SimulatedGANApple, Springenberg_2015_CatGAN, Zhu_2017_CycleConsistentGAN}. The common principle behind all variants of the GAN architecture is to exploit the discriminative power of neural networks in training generative counterparts. GAN models represent a compromise between ease of sampling and perceptual quality of synthesized samples.

Similar to other generative models, the ultimate objective of GANs is to model a target distribution from a limited set of observations. To this end, a basic GAN architecture consists of a cascade of two sub-models, a \emph{generator} $g$ whose objective is to synthesize samples $\hat{\underline{x}}$ indistinguishable from real observations $\underline{x}$, and a \emph{discriminator} $h$ whose objective is to train the generator by estimating a \emph{distance} between the generator's induced distribution $\mathbb{P}_{\hat{\underline{x}}}$ and the target one $\mathbb{P}_{{\underline{x}}}$. In spite of several difficulties in their training process, GAN models proved to be successful in synthesizing plausible, realistic samples while maintaining ease of sampling even for high-dimensional densities \cite{Arjovsky_2017_WassersteinGAN}. Consequently, extensive research has been carried out in order to understand and improve the training process of GAN models \cite{Salimans_2016_ImprovedGAN, Arjovsky_2017_PrincipledMethods, Arjovsky_2017_WassersteinGAN, Mao_2017_LSGAN, Yi_2017_DualGAN}, introduce this model to novel application domains \cite{Isola_2016_Pix2Pix, Zhu_2017_CycleConsistentGAN, Reed_2016_Text2Image, Zhu_2016_iGAN, Li_2016_TextureGAN}, and improve the basic formulation for more advanced learning tasks \cite{Springenberg_2015_CatGAN, Denton_2015_LaplacianPyramids, Wang_2016_S2GAN, Mirza_2014_cGAN, Makhzani_2015_AdversarialAE}.

In this paper, we propose an extension to the basic GAN architecture that targets a more detailed representation of the real data manifold. More specifically, in the proposed architecture, multiple generators cooperate in resembling a target data distribution while maintaining distinct manifolds. All generators are trained using a single adversarial discriminator but are also constrained to distinct distributions via a set of supplementary discriminators. The ultimate objective of the architecture is a set of generative functions learned from a finite set of data samples that jointly model the target data distribution but individually resemble distinct modes in it.

\section{Related work}
\label{sec:realted-work}

The original architecture of a GAN model consisted of a single generator, whose objective is to synthesize data samples that are indistinguishable from real ones, and an adversarial discriminator, which attempts to distinguish between real data samples and generated ones \cite{Goodfellow_2014_GAN, Goodfellow_2016_NIPSTutorial}. Both models are trained simultaneously and in an adversary. When successfully trained, the generator tends to synthesize realistic samples, whereas the discriminator is expected to become maximally confused. Sampling in this architecture consists of sampling a latent variable from a simple distribution (e.g., standard Gaussian) and mapping its value to the sample space using the generator's function. Despite the simplicity of such a process, it is unrestrained over the whole target data manifold. In other words, there is no control on the samples being synthesized by the generator.

In \cite{Radford_2015_DCGAN}, it was observed that the generator tends to structure the latent space, and the complex transformations in the sample space (e.g., face pose) can be controlled via a simple linear interpolation on the latent variable. This feature was exploited in \cite{Chen_2016_InfoGAN} by discretizing a subset of the latent variable attributes in order to control the sampling process. This approach, however, limits the trained generator to a single estimation of the density function whose output can be controlled by conditioning on the selected attributes. In a like manner, Conditional GANs (cGAN) \cite{Mirza_2014_cGAN, Sricharan_2017_SScGAN} controlled this sampling process by conditioning the generator on an observed variable in addition to the latent one. Examples of conditioning variables include text embeddings \cite{Reed_2016_Text2Image}, foreign domain image \cite{Isola_2016_Pix2Pix, Denton_2015_LaplacianPyramids}, or class attributes \cite{Mirza_2014_cGAN}. However, cGAN models require the additional information found in the conditioning variable and are incapable of learning hidden structures of the target distribution in an unsupervised fashion. 

Coupled GAN models \cite{Liu_2016_CoGAN} introduced a multi-generator multi-discriminator architecture whose objective is to learn the joint distribution of multi-domain images from their marginal ones. For a given realization of the latent variable, each generator synthesizes the same sample but in different domains. While this approach eliminates the need for inter-domain knowledge, it requires the complete knowledge of each domain in advance.

Categorical GANs \cite{Springenberg_2015_CatGAN} resemble one of the promising approaches to unravel hidden structures in unlabeled data manifolds. The binary logistic discriminator in traditional GAN models is replaced by a multi-class one whose objective is to maintain high confidence classification scores for the real data samples while being maximally uncertain when provided with synthesized images. As a result, the discriminator becomes capable of clustering the real data samples in a completely unsupervised fashion. Nevertheless, this approach adopts a single generator which is still uncontrolled in its sampling process.

Our approach differs in that its objective is the controlled generation process rather than discrimination. The proposed GAN model attempts to control the sampling process by training several generators that collectively resemble the targeted distribution while maintaining distinct modes among these generators. One can selectively sample from a certain sub-population by sampling from the corresponding generator only.

\section{Model Architecture}
\label{sec:model-architecture}

Let $\underline{x} \in \mathcal{X}$ be the observed data of a continuous random vector $\underline{x}$ from the space $\mathcal{X}$ of observations. The notation $\underline{x} \sim p_{\underline{x}}$ is used to denote the probability density function (pdf) $p_{\underline{x}}$ of $\underline{x}$ of the real data distribution $\mathbb{P}_{\underline{x}}$.

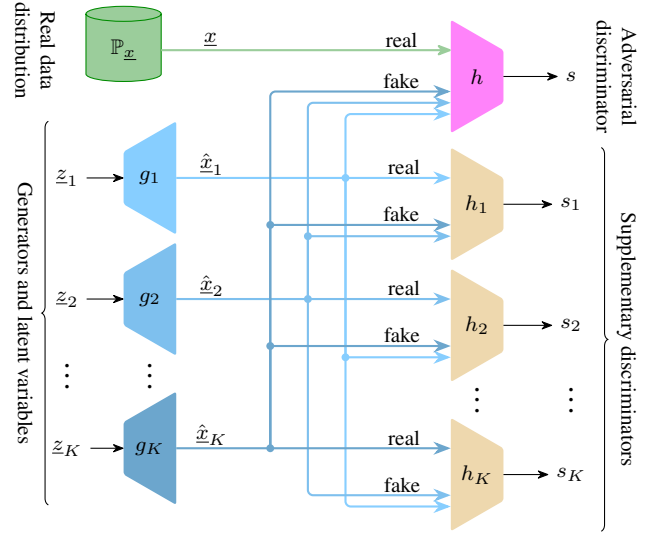
\begin{figure}[!t]
	\centering
\def\zgsep{7.5mm}
\def\gdsep{40mm}
\def\dssep{12.5mm}
\def\ysepa{14mm}
\def\ysepb{7.5mm}
\def\trpw{7.0mm}
\def\trph{7.0mm}
\def\tang{60}
\begin{tikzpicture}[every node/.style={inner sep=0.1mm,outer sep=0mm, font=\footnotesize},
generator/.style={anchor=north,      trapezium, trapezium angle=\tang, minimum width=\trpw, minimum height=\trph, rotate=90, rounded corners=1pt,  fill=SkyBlue1},
suppdiscr/.style={anchor=south west, trapezium, trapezium angle=\tang, minimum width=\trpw, minimum height=\trph, rotate=270, rounded corners=1pt, fill=Wheat2    },
score/.style={draw=none, anchor=center, outer sep=1mm},
latent/.style={draw=none, anchor=center, outer sep=1mm}]
\node (z1) [latent]                          {$\underline{z}_1$}; 
\node (z2) [latent, below=\ysepa of z1.center] {$\underline{z}_2$}; 
\node (z3) [latent, below=\ysepb of z2.center] {\large  $\cvdots$}; 
\node (z4) [latent, below=\ysepb of z3.center] {$\underline{z}_K$}; 

\node (g1) at ($(z1.center) + (\zgsep, 0mm)$) [generator, fill=SkyBlue1] { }; 
\node (g2) at ($(z2.center) + (\zgsep, 0mm)$) [generator, fill=SkyBlue2] { }; 
\node (g3) at (z3.center -| g2.center) [] {\large  $\cvdots$};  
\node (g4) at ($(z4.center) + (\zgsep, 0mm)$) [generator, fill=SkyBlue3] { }; 

\node (d1) at ($(g1.center) + (\gdsep, 0mm)$) [suppdiscr] { };  
\node (d2) at ($(g2.center) + (\gdsep, 0mm)$) [suppdiscr] { };        
\node (d4) at ($(g4.center) + (\gdsep, 0mm)$) [suppdiscr] { };  
\node (d3) at ($(d2.center)!0.5!(d4.center)$) [] {\large  $\cvdots$};         
\node (d0) at ($(d1.south west) + (0mm, 17mm)$) [suppdiscr, fill=Orchid1] {};         

\node (s0) at ($(d0.center) + (\dssep, 0mm)$) [score] {${s}$};
\node (s1) at ($(d1.center) + (\dssep, 0mm)$) [score] {${s}_1$};
\node (s2) at ($(d2.center) + (\dssep, 0mm)$) [score] {${s}_2$};
\node (s3) at (d3.center -| s2.center)       {\large  $\cvdots$};
\node (s4) at ($(d4.center) + (\dssep, 0mm)$) [score] {${s}_K$};

\node at (g1.center) {$g_1$}; 
\node at (g2.center) {$g_2$}; 
\node at (g4.center) {$g_K$}; 

\node at (d0.center) {$h$}; 
\node at (d1.center) {$h_1$}; 
\node at (d2.center) {$h_2$}; 
\node at (d4.center) {$h_K$}; 

\node (g0) at (g1.north |- d0.south west) [anchor=center, cylinder, shape border rotate=90, inner sep=1mm, minimum height=10mm, minimum width=10mm, fill=DarkSeaGreen3, draw=Green4] {};
\node at (g0.center) [anchor=center] {$\mathbb{P}_{\underline{x}}$};

\draw [arrows={ - Stealth[round]}, thick, DarkSeaGreen3] (g0.east) -- (d0.south west) 
node [pos=0.17, above=0.35mm, black] {$\underline{x}$} 
node [pos=0.83, above=0.35mm, black] {real};

\draw [arrows={ - Stealth[round]}] (d0.north) -- (s0.west);        
\draw [arrows={ - Stealth[round]}] (z1.east) -- (g1.north); \draw [arrows={ - Stealth[round]}] (d1.north) -- (s1.west);        
\draw [arrows={ - Stealth[round]}] (z2.east) -- (g2.north); \draw [arrows={ - Stealth[round]}] (d2.north) -- (s2.west);        
\draw [arrows={ - Stealth[round]}] (z4.east) -- (g4.north); \draw [arrows={ - Stealth[round]}] (d4.north) -- (s4.west);        

\node (g1split) at ($(d1.south west) - (14mm, 0)$) [anchor=center, inner sep=0.37mm, outer sep=0mm, circle, fill=SkyBlue1] {};
\draw [arrows={- Stealth[]}, rounded corners=2pt, SkyBlue1, thick] (g1split.center) |- ($(d0.south east) - (0mm, 1.50mm)$);
\draw [arrows={- Stealth[]}, rounded corners=2pt, SkyBlue1, thick] (g1split.center) |- ($(d4.south east) - (0mm, 0.75mm)$);
\draw [arrows={- Stealth[]}, rounded corners=0pt, SkyBlue1, thick] (g1split.center) |- ($(d2.south east) - (0mm, 0.75mm)$)
node [pos=0.5, anchor=center, inner sep=0.37mm, outer sep=0mm, circle, fill=SkyBlue1] {};
\draw [arrows={ - Stealth[]}, SkyBlue1, thick] (g1.south) -- (d1.south west) 
node [pos=0.13, above=0.35mm, black] {$\hat{\underline{x}}_1$} 
node [pos=0.83, above=0.35mm, black] {real};

\node (g2split) at ($(d2.south west) - (19mm, 0)$) [anchor=center, inner sep=0.37mm, outer sep=0mm, circle, fill=SkyBlue2] {};
\draw [arrows={- Stealth[]}, rounded corners=2pt, SkyBlue2, thick] (g2split.center) |- ($(d0.south east) - (0mm, 0.0mm)$);
\draw [arrows={- Stealth[]}, rounded corners=2pt, SkyBlue2, thick] (g2split.center) |- ($(d4.south east) + (0mm, 0.75mm)$)
node [pos=0.83, above=0.35mm, black] {fake};
\draw [arrows={- Stealth[]}, rounded corners=0pt, SkyBlue2, thick] (g2split.center) |- ($(d1.south east) - (0mm, 0.75mm)$)
node [pos=0.5, anchor=center, inner sep=0.37mm, outer sep=0mm, circle, fill=SkyBlue2] {};
\draw [arrows={ - Stealth[]}, SkyBlue2, thick] (g2.south) -- (d2.south west) 
node [pos=0.13, above=0.35mm, black] {$\hat{\underline{x}}_2$} 
node [pos=0.83, above=0.35mm, black] {real};

\node (g4split) at ($(d4.south west) - (24mm, 0)$) [anchor=center, inner sep=0.37mm, outer sep=0mm, circle, fill=SkyBlue3] {};
\draw [arrows={- Stealth[]}, rounded corners=2pt, SkyBlue3, thick] (g4split.center) |- ($(d0.south east) + (0mm, 1.50mm)$)
node [pos=0.863, above=0.35mm, black] {fake};
\draw [arrows={- Stealth[]}, rounded corners=0pt, SkyBlue3, thick] (g4split.center) |- ($(d1.south east) + (0mm, 0.75mm)$)
node [pos=0.5, anchor=center, inner sep=0.37mm, outer sep=0mm, circle, fill=SkyBlue3] {}
node [pos=0.863, above=0.35mm, black] {fake};
\draw [arrows={- Stealth[]}, rounded corners=0pt, SkyBlue3, thick] (g4split.center) |- ($(d2.south east) + (0mm, 0.75mm)$)
node [pos=0.5, anchor=center, inner sep=0.37mm, outer sep=0mm, circle, fill=SkyBlue3] {}
node [pos=0.865, above=0.35mm, black] {fake};
\draw [arrows={ - Stealth[]}, SkyBlue3, thick] (g4.south) -- (d4.south west) 
node [pos=0.13, above=0.35mm, black] {$\hat{\underline{x}}_K$} 
node [pos=0.83, above=0.35mm, black] {real};

\draw [decorate, decoration={brace, aspect=0.5, raise=20mm}] (d1.bottom left corner) -- (d4.bottom right corner)
node[pos=0.5, right, rotate=-90, anchor=south, yshift=22mm] {Supplementary discriminators};
\draw [decorate, decoration={brace, aspect=0.5, raise=17mm}] (g4.bottom left corner) -- (g1.bottom right corner)
node[pos=0.5, right, rotate=-90, anchor=north, yshift=-19mm] {Generators and latent variables};
\node at (s0.east) [rotate=-90, anchor=south, yshift=1.5mm, align=center] {Adversarial \\ discriminator};
\node at (g0.west) [rotate=-90, anchor=north, yshift=-4.0mm, align=center] {Real data \\ distribution};

\end{tikzpicture}
\caption{{\small A diagram of the proposed multi-generator GAN architecture. Multiple inputs to each discriminator is essentially aggregation of sample from different distributions into a single batch.}}
\label{fig:model-architecture}
	\vspace{-5.0mm}
\end{figure}

The proposed architecture is depicted in Fig. \ref{fig:model-architecture} and consists of $K$ generative models (generators) $\{g_k\}_{k=1}^{K}$, each of which maps a latent variable $\underline{z}_k\sim p_{\underline{z}}$ to a synthesized sample $\smash{\hat{\underline{x}}_k = g_k(\underline{z}_k)}\sim p_k$ and induces a continuous\footnote{For simplicity, and for illustrative purposes, we shall assume that all distributions are continuous with non-disjoint support.} distribution $\mathbb{P}_k$. All synthesized samples $\hat{\underline{x}}_k$ belong to the same space $\mathcal{X}$. Let $p_{\underline{x}}$ be the density function of the real data distribution $\smash{\mathbb{P}}_{\underline{x}}$. Furthermore, let $\hat{\underline{x}}$ be a random vector equi-probably sampled from $\{g_k\}_{k=1}^{K}$. This results in a mixture pdf $p_{\hat{\underline{x}}}$ defined as
\begin{equation}
p_{\hat{\underline{x}}}(\underline{x}) =\; \frac{1}{K} \sum_{k=1}^K p_k(\underline{x})
\label{eq:mixture-model}
\end{equation}%
We further introduce $K$ binary logistic models (discriminators) $\{h_k\}_{k=1}^{K}$ each of which maps samples from the input data space $\hat{\underline{x}}\sim\mathcal{X}$ to a probability $h_k: \mathcal{X} \rightarrow [0, 1]$ where $h_k(\hat{\underline{x}})$ resembles the probability that $\hat{\underline{x}}$ is sampled from the $k^\text{th}$ model $g_k$ as opposed to all other components in the mixture. The opposite probability that $\hat{\underline{x}}$ is equi-probably sampled from the remaining set of generators $g_j,\,j\neq k$ is
\begin{equation}
p_{\bar{k}}(\underline{x}) = \frac{1}{K-1} \sum_{\substack{j=1\\j\neq k}}^{K} p_j(\underline{x})
\end{equation}
which also gives another formulation for the mixture pdf
\begin{equation}
p_{\hat{\underline{x}}}(\underline{x}) =\; \frac{1}{K} \sum_{k=1}^K p_{\bar{k}}(\underline{x})
\label{eq:complement-mixture-model}
\end{equation}
The loss function of each of the $k^\text{th}$ discriminator becomes
{\small
\begin{align*}
\mathcal{L}_{h_k}  
& =                                    \mathbb{E}_{\underline{z} \sim p_{\underline{z}}         }             \log\left(     h_k(g_k(\underline{z})) \right) + 
    \sum_{\substack{j=1\\j\neq k}}^{K} \mathbb{E}_{\underline{z} \sim p_{\underline{z}}         }             \log\left( 1 - h_k(g_j(\underline{z})) \right) \\
& =                                    \mathbb{E}_{\underline{x} \sim p_k         }             \log\left(     h_k(    \underline{x})  \right) + 
    \sum_{\substack{j=1\\j\neq k}}^{K} \mathbb{E}_{\underline{x} \sim p_j         }             \log\left( 1 - h_k(    \underline{x})  \right) \\
& =                                    \mathbb{E}_{\underline{x} \sim p_k         }             \log\left(     h_k(    \underline{x})  \right) + 
                                       \mathbb{E}_{\underline{x} \sim p_{\bar{k}} } \left[(K-1) \log\left( 1 - h_k(    \underline{x})  \right) \right] \\
\end{align*}}%

Moreover, an additional adversarial discriminator $h$ distinguishes samples of the real mixture distribution $\mathbb{P}_{\underline{x}}$ from those of the generators' mixture $\mathbb{P}_{\hat{\underline{x}}}$. The loss function of this discriminator is similar to the traditional GAN loss 
{\small \begin{align*}
	\mathcal{L}_{h}  
	& =\mathbb{E}_{\underline{x} \sim p_{     \underline{x}}}\log\left(     h(\underline{x}) \right) + 
	\mathbb{E}_{\underline{x} \sim p_{\hat{\underline{x}}}}\log\left( 1 - h(\underline{x}) \right) \\
	\end{align*}}%

Let $V$ be the value function of the \emph{mini-minimax} game defined as
{\small \begin{align}
	\min_{\{g_k\}} \min_{\{h_k\}} \max_{h} V = \mathcal{L}_{h} - \sum_{k=1}^{K} \mathcal{L}_{h_k} 
	\end{align}}%
The adversarial GAN game in this case becomes a \emph{minimax} game between two teams. The first team consists of a single player (the adversarial discriminator $h$) whose objective is to \emph{maximize} the value function. The second team consists of all generators $\{g_k\}_{k=1}^{K}$ and supplementary discriminators $\{h_k\}_{k=1}^{K}$ where each player in this team is tasked with \emph{minimizing} the same value function. Each model optimizes the value function with respect to its own parameters (assuming all other models are fixed).

In the non-parametric limit, and assuming each model can train to completion, the resulting discriminators reach their global optimum at (proof can be found in \cite{Goodfellow_2014_GAN})
{\small \begin{equation*}
h_k^\star(\underline{x}) =\, \frac{p_k(\underline{x})}{p_k(\underline{x}) + (K-1) \cdot p_{\bar{k}}(\underline{x})} =\, \frac{p_k(\underline{x})}{K \cdot p_{\hat{\underline{x}}}(\underline{x})}
\end{equation*}}%
and
{\small \begin{equation*}
h^\star(\underline{x})   =\, \frac{p_{{\underline{x}}}(\underline{x})}{p_{\hat{\underline{x}}}(\underline{x})\; +\; p_{{\underline{x}}}(\underline{x})}
\end{equation*}}%
Substituting these optimal values in the value function $V$ results in 
{\small \begin{alignat*}{3}
V && =\; &&&                       \mathbb{E}_{\underline{x} \sim p_{\underline{x}}      }        \log\left(     h^\star  (\underline{x}) \right) 
           +                       \mathbb{E}_{\underline{x} \sim p_{\hat{\underline{x}}}}        \log\left( 1 - h^\star  (\underline{x}) \right)  \\
  && -\; &&& \sum_{k=1}^{K} \left( \mathbb{E}_{\underline{x} \sim p_k                    }        \log\left(     h_k^\star(\underline{x}) \right) +
                                   \mathbb{E}_{\underline{x} \sim p_{\bar{k}}            }        \log\left( 1 - h_k^\star(\underline{x}) \right) \right) \\
  && =\; &&&                       \mathbb{E}_{\underline{x} \sim p_{\underline{x}}      }        \log\left(\frac{p_{{\underline{x}}}(\underline{x})}{p_{\hat{\underline{x}}}(\underline{x})\; +\; p_{{\underline{x}}}(\underline{x})}     \right) \\ 
  && +\; &&&                       \mathbb{E}_{\underline{x} \sim p_{\hat{\underline{x}}}}        \log\left(\frac{p_{\hat{\underline{x}}}(\underline{x})}{p_{\hat{\underline{x}}}(\underline{x})\; +\; p_{{\underline{x}}}(\underline{x})} \right) \\
  && -\; &&& \sum_{k=1}^{K}        \mathbb{E}_{\underline{x} \sim p_k                    }  \left[\log\left(\frac{p_k                    (\underline{x})}{p_{\hat{\underline{x}}}(\underline{x})}  \right) + \log\left(\frac{1}{K}\right)                      \right] \\
  && -\; &&& \sum_{k=1}^{K}        \mathbb{E}_{\underline{x} \sim p_{\bar{k}}            }  \left[\log\left(\frac{p_{\bar{k}            } (\underline{x})}{p_{\hat{\underline{x}}}(\underline{x})} \right) + \log\left(\frac{K-1}{K}\right)                    \right] \\
\end{alignat*}\vspace{-10mm}}%

All terms in the previous equation resemble different, non-symmetric Kullback-Leibler (KL) divergences resulting in 
{\small \begin{alignat}{3}
V && =\; &&&                \KLD \left( \mathbb{P}_{\underline{x}}        \;\|\;  \mathbb{P}_{\hat{\underline{x}}} \plus\; \mathbb{P}_{{\underline{x}}}    \right) + 
                            \KLD \left( \mathbb{P}_{\hat{\underline{x}}}  \;\|\;  \mathbb{P}_{\hat{\underline{x}}} \plus\; \mathbb{P}_{{\underline{x}}}    \right) \nonumber \\ 
  && -\; &&& \sum_{k=1}^{K} \KLD \left( \mathbb{P}_k                      \;\|\;  \mathbb{P}_{\hat{\underline{x}}}                                         \right) - 
             \sum_{k=1}^{K} \KLD \left( \mathbb{P}_{\bar{k}}              \;\|\;  \mathbb{P}_{\hat{\underline{x}}}                                         \right) \nonumber \\
  && -\; &&& K\cdot\log\left(\frac{K-1}{K^2}\right) 
\label{eq:generator-value-function-kl}
\end{alignat}}%
It can be further simplified to symmetric Jensen-Shannon (JS) divergences
{\small \begin{alignat}{3}
V  && =\; &&& 2\cdot\JSD \left( \mathbb{P}_{\hat{\underline{x}}},  \mathbb{P}_{{\underline{x}}}   \right) \nonumber - K\cdot \JSD \left( \mathbb{P}_{{1}}, \dots, \mathbb{P}_{{K}}  \right) \nonumber \\
   && -\; &&&  K\cdot\JSD \left( \mathbb{P}_{\bar{1}}, \dots, \mathbb{P}_{\bar{K}} \right) - K\cdot\log\frac{K-1}{K^2} - \log 4
\label{eq:generator-value-function}
\end{alignat}}%
where the $\pi$-generalized JS divergence for a finite mixture of distributions is defined as \cite{Lin_1991_ShannonDivergences}
{\small \begin{equation}
    \JSD^{\pi_1, \dots, \pi_K}(\mathbb{P}_{1}, \dots, \mathbb{P}_K) =\; \sum_{k=1}^{K}\pi_k\cdot\KLD\left(\mathbb{P}_k\;\Big\|\;\sum_{j=1}^{K}\pi_j\cdot\mathbb{P}_j\right)
    \label{eq:jensen-shannon-mixture}
\end{equation}}%
where $\pi_1,\dots,\pi_K$ are the distributions' weights and are assumed in this work to be $1/K$. Eq. \ref{eq:jensen-shannon-mixture} is applicable to the two distributions $\mathbb{P}_{{\underline{x}}}$ and $\mathbb{P}_{\hat{\underline{x}}}$ by defining an average distribution $\mathbb{P}_A = (\mathbb{P}_{\hat{\underline{x}}} + \mathbb{P}_{{\underline{x}}}) / 2$. It is also applicable within the mixture model $\mathbb{P}_{\hat{\underline{x}}}$ using Eq. \ref{eq:mixture-model} and \ref{eq:complement-mixture-model}.
Minimizing Eq. \ref{eq:generator-value-function} is equivalent to minimizing the \emph{difference} between the real data distribution and the estimated one while maximizing the dissimilarity between the components of the latter. This, in turn, induces each generator in the model to maintain distinct observational claims of the real data distribution.

Eq. \ref{eq:generator-value-function-kl} represents a very interesting objective function. The first two terms resemble the symmetric JS divergence between the real distribution and the estimated mixture which is non-negative and retains its minimum when $\mathbb{P}_{\underline{x}} = \mathbb{P}_{\hat{\underline{x}}}$. This term is the common measure in traditional GAN models \cite{Goodfellow_2014_GAN}. The third KL-divergence prevents any of the component models $g_k$ from prevalence (representing the whole mixture all by itself), whereas the last term prevents the same component from vanishing. 

In practice, each model (either a discriminator or a generator) is a parametric Multi-Layer Perceptron (MLP). This, of course, limits the family of density functions representable by each generator. Nevertheless, each one retains its own parameters, and as a result, the estimated distribution represents a heterogeneous mixture model. Training is straightforward and, similar to traditional GANs, consists of alternating gradient-based parameter updates. Gradient-ascent is utilized by the adversarial discriminator while the supplementary discriminators and the generators use gradient-descent.

\section{Experiments and results}
\label{sec:experiments-and-results}

\subsection{Experimental setup}
\label{subsec:experimental-setup}
We test and evaluate the proposed model on the MNIST dataset \cite{LeCun_1998_LeNet} and with the two-model case\footnote{We delegate the many model $K>2$ case to a future work. Our initial experiments show that the derived value function is unbalanced (i.e., divergences are unequally bounded and weighted) which results in degradation of the perceptual quality as the number of models increase.} $K=2$. In each test case, we select images from a two-digit combination (e.g, $\{0,\,1\}$) and train the randomly initialized model on the unlabeled dataset. We expect that each of the two generators will synthesize images of a unique digit. 

The latent random vector is uniformly distributed $\underline{z}\sim \mathcal{U}(\minus\,1, 1)$ in a 100-dimensional space $\underline{z}\in \mathbb{R}^{100}$ and it seeds two generators $g_1$ and $g_2$ where each is a 2-layer, feed-forward, fully connected neural network. Both the observed variable and the synthesized one belong to the same space $\underline{x},\hat{\underline{x}} \in \mathbb{R}^{784}$ and both are used to train the adversarial discriminator\footnote{A single binary discriminator for $g_1$ petted against $g_2$.} $h$. We use a single supplementary discriminator $h_1$ whose input is a batch of the synthesized images $\hat{\underline{x}}$. The two discriminators $h$ and $h_1$ are identically structured and have exactly the inverse architecture of the generators (except for the output layer). The output layer of each model uses logistic sigmoidal activation whereas its hidden layers use rectified linear units \cite{Nair_2010_ReLU_RBM}. The whole model is 4-layer deep in its longest path and comprises 0.8M trainable parameters. Finally, all models are trained using the Adam optimizer \cite{Kingma_2014_ADAM} with a learning rate of $10^{-3}$ and each sub-model gains a single parameter-update step in each iteration.

We highlight that instead of adhering to the aforementioned theoretical analysis in which training alternates between gradient-descent and -ascent steps, we rather flip the desired output $y\leftarrow 1-y$ and use only gradient-descent. This technique has been proposed in several previous works \cite{Goodfellow_2016_NIPSTutorial, Mao_2017_LSGAN} with conceivable reasoning\footnote{More recent studies proved that such an approach indeed eliminates the vanishing gradient problem but rather introduces critical sources of instability in the training process \cite{Arjovsky_2017_PrincipledMethods}.} but rather violates the presented analytical study of the architecture since each sub-model retains a different objective function.

\begin{figure}[t]
	\begin{tikzpicture}[every node/.style={inner sep=0,outer sep=0}]
	\node (a) {\includegraphics[width=\linewidth, totalheight=0.13\linewidth]{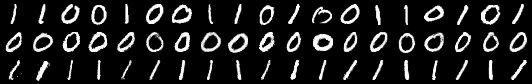}};
	\node (a) [below = 2mm of a.south, anchor=north] {\includegraphics[width=\linewidth, totalheight=0.13\linewidth]{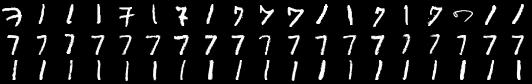}};
	\node (a) [below = 2mm of a.south, anchor=north] {\includegraphics[width=\linewidth, totalheight=0.13\linewidth]{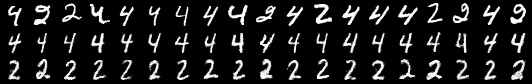}};
	\node (a) [below = 2mm of a.south, anchor=north] {\includegraphics[width=\linewidth, totalheight=0.13\linewidth]{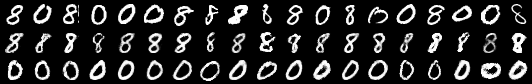}};
	\node (a) [below = 2mm of a.south, anchor=north] {\includegraphics[width=\linewidth, totalheight=0.13\linewidth]{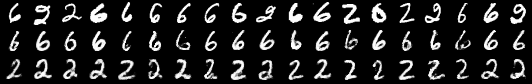}};
	\end{tikzpicture}
	\caption{Two-digit experiments on the proposed architecture using the MNIST dataset. The top-most row of each experiment shows
		a sample of the real images whereas the two lower rows depict synthesized samples from each generator.}
	\label{fig:two-digit-success}
	\vspace{-2.5mm}
\end{figure}
\begin{figure}[!ht]
	\begin{tikzpicture}[every node/.style={inner sep=0,outer sep=0}]
	\node (a) {\includegraphics[width=\linewidth, totalheight=0.13\linewidth]{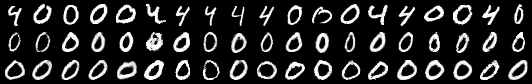}};
	\node (a) [below = 2mm of a.south, anchor=north] {\includegraphics[width=\linewidth, totalheight=0.13\linewidth]{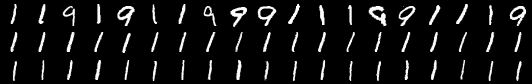}};
	\end{tikzpicture}
	\caption{Examples of failure cases.}
	\label{fig:two-digit-failure}
	\vspace{-2.5mm}
\end{figure}

\subsection{Results and discussion}
\label{subsec:dicussion}

Fig. \ref{fig:two-digit-success} shows the output of each generator in various experiments (each with a distinct two-digit combination). As observed, the generators produce samples of moderate to high perceptual quality but each one synthesizes a different digit from the given combination. On the other hand, few failure cases were observed and are shown in Fig. \ref{fig:two-digit-failure}. In these experiments, we still observe that the generators tend to produce, to some extent, different representations of the same digit rather than distinct digits.

Training adversarial networks, in general, is known to be unstable \cite{Arjovsky_2017_PrincipledMethods}, and the alternating gradient-based parameter updates (rather than training to completion) between the two sub-models in the original model is one of the sources for their instability \cite{Arjovsky_2017_PrincipledMethods}. In the proposed architecture, these alternating parameter updates are extended to a larger number of sub-models and, as a result, expected to amplify this instable training behavior. However, we observed in our experiments that the generators maintain distinct manifolds even in the early training steps and only collapse to the same digit when over-trained.

Furthermore, the objective of the supplementary discriminators is to provide a simple means for exploiting the powerful back-propagation algorithm in pushing the generators' distributions apart from each other. This seems to be valid as long as these models are weak classifiers. If they are provided with enough training capacity, they begin to deliver their discrimination task all by themselves and their driving force on the generators' distributions disappears. As a results, the generators tend to collapse to the same mode which is highly-rewarded by the adversarial discriminator rather than adhering to distinct classes. In fact, the precise amount of training, even for traditional GAN models, is a challenging hyper-parameter to optimize for reasons explained in \cite{Arjovsky_2017_PrincipledMethods}.

%

\section{Conclusion}
\label{sec:conclusion}

We proposed a more detailed representation learning model based on a multi-generator extension to the GAN architecture. The generators tend to cooperatively resemble the target data distribution but each generator tends to produce only a distinct mode. This architecture provides selective sampling from a certain cluster of the target without the need of any labeling information.


 {\small\vspace{-1mm}
 \bibliographystyle{./bib/IEEEtranKarim}
 \bibliography{./bib/IEEEabrv,./bib/references}}


\end{document}